\documentclass[conference]{IEEEtran}
\IEEEoverridecommandlockouts
\usepackage{cite}
\usepackage{amsmath,amssymb,amsfonts}
\usepackage{algorithmic}
\usepackage{graphicx}
\usepackage{textcomp}
\usepackage{xcolor}
\usepackage[utf8]{inputenc}

\def\BibTeX{{\rm B\kern-.05em{\sc i\kern-.025em b}\kern-.08em
    T\kern-.1667em\lower.7ex\hbox{E}\kern-.125emX}}
        
\usepackage{fancyhdr}
\usepackage{caption}

\newcommand{\suparrow}{{\uparrow}}
\newcommand{\sdownarrow}{{\downarrow}}
\newcommand\allbold[1]{{\boldmath\textbf{#1}}}

\newcommand{\lhcmt}[1]{#1}
\newcommand{\lhcorr}[2]{#2}

\newcommand{\bigbold}[1]{{\fontsize{8}{12}\allbold{#1}}}

\begin{document}

\title{Slices of Attention in Asynchronous Video Job Interviews\\
}

\author{\IEEEauthorblockN{Léo Hemamou\IEEEauthorrefmark{1}\IEEEauthorrefmark{2}\IEEEauthorrefmark{3}, Ghazi Felhi\IEEEauthorrefmark{1}, Jean-Claude Martin \IEEEauthorrefmark{2} and Chloé Clavel\IEEEauthorrefmark{3}}
\IEEEauthorblockA{\IEEEauthorrefmark{1}EASYRECRUE, 38 Rue du Sentier, 75002 Paris, France \\
Email: \{l.hemamou,g.felhi\}@easyrecrue.com},
\IEEEauthorblockA{\IEEEauthorrefmark{2}LIMSI, CNRS, Paris-Sud University, Paris-Saclay University / F-91405 Orsay, France\\
Email: Jean-Claude.Martin@limsi.fr}\IEEEauthorblockA{\IEEEauthorrefmark{3}LTCI, Télécom ParisTech, Paris-Saclay University / F-75013 Paris, France\\
Email: chloe.clavel@telecom-paristech.fr}}


\maketitle
\thispagestyle{fancy}

\begin{abstract}
The impact of non verbal behaviour in a hiring decision remains an open question. Investigating this question is important, as it could provide a better understanding on how to train candidates for job interviews and make recruiters be aware of influential non verbal behaviour. This research has recently been accelerated due to the development of tools for the automatic analysis of social signals (facial expression detection, speech processing, etc), and the emergence of machine learning methods. However, these studies are still mainly based on hand engineered features, which imposes a limit to the discovery of influential social signals. On the other side, deep learning methods are a promising tool to discover complex patterns without the necessity of feature engineering. In this paper, we focus on studying influential non verbal social signals in asynchronous job video interviews that are discovered by deep learning methods. We use a previously published deep learning system that aims at inferring the hirability of a candidate with regard to a sequence of interview questions. One particularity of this system is the use of attention mechanisms, which aim at identifying the relevant parts of an answer. Thus, information at a fine-grained temporal level could be extracted using global (at the interview level) annotations on hirability. While most of the deep learning systems use attention mechanisms to offer a quick visualization of slices when a rise of attention occurs, we perform an in-depth analysis to understand what happens during these moments. First, we propose a methodology to automatically extract slices where there is a rise of attention (\textit{attention slices}). Second, we study the content of\textit{ attention slices} by comparing them with randomly sampled slices. Finally, we show that they bear significantly more information for hirability than randomly sampled slices, and that such information is related to visual cues associated with anxiety and turn taking.

\end{abstract}


\section{Introduction}

The procedure of personnel selection includes gathering data about the potential candidates, for example, in a job interview \cite{2012TheSelection}. Research in Affective Computing can be helpful in many ways with respect to job interviews, for example virtual recruiters can help candidates train their social skills and rehearse \cite{Hoque2013MACHb}. This automatic processing can help recruiters assess candidates. Additionally, it can help researchers and recruiters understand the evaluation process done by recruiters when assessing a candidate\lhcorr{, with a potential for preventing biases in the hiring process}{}.
Initially conducted face to face or via phone, job interviews are now often done by online video conferencing systems or by asynchronous video recordings. An asynchronous video interview is an emergent tool now offered by several companies responding to the needs of initial assessment in personnel selection. The procedure is as follows: the candidate connects to a web platform and answers a sequence of questions predefined by the recruiter while recording a video of himself with his webcam, smartphone or tablet. Later, recruiters connect to the same platform, watch the candidate's answers, rate the answers and then decide whether they want to invite the candidate to a face-to-face interview.

Researchers are already developing systems for automatically predicting hirability based on non verbal cues of candidates in asynchronous video interviews\cite{Chen2017,Nambiar2017AutomaticInterviews}. In this context and in addition to new legislative constraints (General Data Protection Regulation), such automatic systems require interpretability and transparency. With these systems, candidates will be able to improve their non verbal behavioral strategy, and recruiters will be able to assess these decision support models. These models could even help recruiters understand their own biases. 

Classical approaches in social computing consist of building machine learning models and interpreting the importance of the features \cite{Naim2018AutomatedPerformance,RaoS.B2017AutomaticStudy,Wortwein2015MultimodalAssessment}.\lhcorr{Such methodologies are efficient and highly interpretable, but have multiple drawbacks. Among them, is the impossibility to take into account temporality (eg. the sequentiality of questions during an interview), and the inability to bring forward and stress the effect of unexpected short and influential social cues.}{ These approaches are not able to bring forward and stress the effect of unexpected and influential social cues.} 
We previously proposed a deep learning model trained only with recruiter's decision and videos \cite{Hemamou2019HireNetInterviews}. Our model is able to consider temporality and influential slices of video-based asynchronous interviews, due to the use of an attention mechanism running on top of a recurrent neural network. A sequence of features is processed by a recurrent neural network, and the attention mechanism aims at learning a different weight for each time step to enhance the performance of the classification task. Overall, such techniques could be useful for understanding human behaviour, as they aim to separate task relevant time steps in a sequence from irrelevant time steps \cite{Yu2017TemporallyContent}. However, research in this area is limited to studying and highlighting just a few examples of peaks of attention curves \cite{Yu2017TemporallyContent,Martins2016FromClassification}. Hence, a consistent validation is needed in order to ascertain the usefulness of the system's output for predicting hirability as rated by recruiters. 

With this in mind, this article describes three experiments we conducted to understand whether slices of video interviews highlighted by the attention model do carry information that is useful to recruiters. In section \ref{Q1}, we propose a methodology to  automatically extract thin slices where attention values are high. In section \ref{Q2}, we test whether the non verbal behavior occurring during these slices is different from behavior occurring in randomly picked slices. Finally, in section \ref{Q3}, we evaluate whether the extracted slices are more informative with regards to the hirability of a candidate.

\begin{figure}[t]
\centerline{\includegraphics[width=1.\columnwidth]{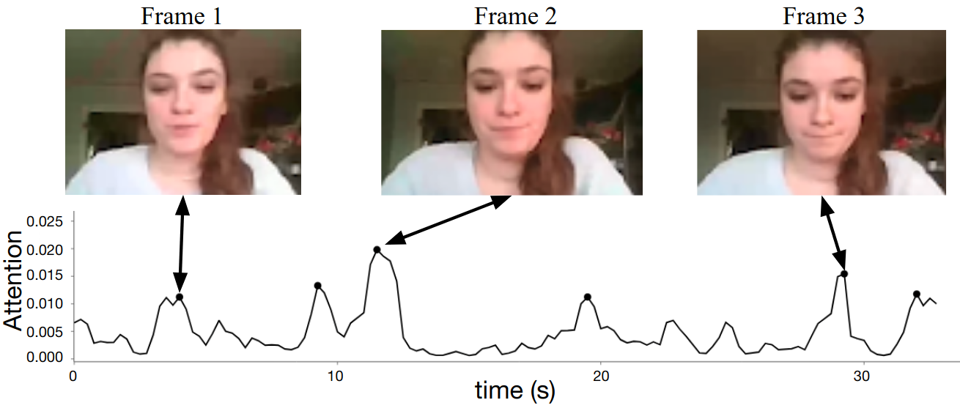}}
\caption{Example of attention curve and salient moments detected with peaks in HireNet.}
\label{figExampleAtt}
\end{figure}

\section{Related Works}
\subsection{Job interview and Non Verbal Behaviour }

\lhcorr{Numerous studies have investigated the effects of non verbal behavior in job interviews.}{Non verbal} visual and audio cues have been studied in order to predict interview performance \cite{FORBES1980NonverbalInterviews}, anxiety\cite{Feiler2016BehavioralAnxiety}, personality of the candidates \cite{Degroot2009CanInterviews} or deception \cite{Schneider2015CuesInterview}. Numerous visual cues such as physical attractiveness, hand gestures, smiling, eye contact, nodding, head movement, body orientation, facial cues, leg movements have been used throughout experiments. For example torso movement, face touching, leg fidgeting, \cite{Feiler2016BehavioralAnxiety} neutral expression and less smiling have been found \cite{Gifford1985NonverbalJudgments.} to negatively correlate to interview performance, whereas eye contact, hand gesture\cite{Feiler2016BehavioralAnxiety}, head movement \cite{Schneider2015CuesInterview} correlates positively with interview performance.
Moreover, these cues could have a different impact on interviewer evaluations depending on interview structure, job position (blue collar vs white collar) or settings of the interview (such as telephone\lhcorr{, instant-message software}{}, computer-mediated video chat and asynchronous video interview)\cite{Frauendorfer2015TheInterview}. 

Putting aside efficiency, annotating every segment of a video is time consuming. 
One common approach to deal with the task of annotations is to only annotate a part of the job interview. In fact, it has been shown that, using only a short amount of information,  people can infer correctly personal characteristics, traits or states of an individual\cite{Murphy2015ReliabilityInteractions,Carney2007AImpressions}. This approach is called thin slice analysis and has already been used in social interactions study \cite{Murphy2015ReliabilityInteractions}, first impressions\cite{Carney2007AImpressions}, public speaking\cite{Chollet2017AssessingBehavior}, or job interviews\cite{Nguyen2015IMinute}. Another advantage of this method is that it highlights brief, non verbal behavior with respect to perceived impressions. Nonetheless, the duration and sampling strategy for thin slices remain an open question. Previous studies focus on sampling thin slices randomly\cite{Chollet2017AssessingBehavior}, using the structure of the job interview (slices based on questions and answers) \cite{Nguyen2015IMinute}, or at the beginning and end of the interactions\cite{Degroot2009CanInterviews}. 
Automatic methods based on social signal processing could give way to a better selection of thin slices and their duration, by selecting regions that carry more information.

\subsection{Automatic methods for understanding human behavior in job interviews}

\lhcorr{Whether on software or on hardware, various advances have been made in the recent years when it comes to social signal analysis. Most of these advances are meant to reduce time spent in manually coding behavioral cues.}{Recent advances drastically reduce time spent in manually coding behavioral cues.} Tools are now available to automatically code vocal \cite{Eyben2016TheComputing} or visual \cite{Baltrusaitis2018OpenFaceToolkit} cues. Recent studies use social signal processing and machine learning to understand the links between non verbal cues and hirability. These studies have been applied to different job interview settings: face to face interviews \cite{Nguyen2014HireBehavior,Nguyen2015IMinute}, asynchronous video interviews \cite{Chen2017} and computer-mediated video chat \cite{RaoS.B2017AutomaticStudy}.

Among investigated traits in job interviews (communication skills\cite{RaoS.B2017AutomaticStudy}, personality \cite{Chen2017}, etc), hirability remains the most studied one. Usually, two methods are used to understand which extracted features are important for hirability: correlation analysis and feature importance analysis conducted on a trained machine learning model. 
However, feature importance analysis is highly dependent on a machine learning pipeline. To the best of our knowledge, only traditional machine learning (SVM, Lasso, Ridge, etc) has been used so far. In a previous work, we investigated the use of deep learning techniques, and established their superiority in terms of predictive capabilities \cite{Hemamou2019HireNetInterviews}. However no in-depth analysis about the feedback returned by the model was conducted.

\subsection{Neural Networks and explainability}

Neural networks are able to find more statistical patterns than traditional machine learning methods such as SVMs, logistic regressions or Random Forests. 
Moreover, specific architectures such as recurrent neural networks allow for modeling temporality by managing sequences. However, the freedom they have to construct intermediate representation comes at the cost of an extreme opacity. This opacity hinders their usability for critical applications such as healthcare, justice, or human resources.
Therefore several researchers have tried to propose methods to better explain these networks. First, the visualization of hidden states has been explored to better understand intermediate representations automatically built by the networks, especially in computer vision \cite{Yosinski2015UnderstandingVisualization}.

A second approach, called knowledge distillation, consists of learning an interpretable model from an already trained complex neural networks \cite{Liu2019ImprovingDistillation}. A third method consists of explaining predictions for specific instances (as opposed to explaining the whole model). Some of the attempts build a local boundary for these predictions \cite{Ribeiro2016WhyClassifier}, using sensitivity methods or analyzing integrated gradients for image analysis \cite{Sundararajan2017AxiomaticNetworks}. Finally, attention mechanisms have recently gained popularity for enhancing performance and interpretability. Specifically in the Social Computing area, the use of attention mechanisms for rapport detection \cite{Yu2017TemporallyContent} or for the evaluation of job interview performance in asynchronous video interviews \cite{Hemamou2019HireNetInterviews} has been proposed to \lhcorr{highlight important slices. The latter approach is very promising as it could potentially be used to}{} extract fine grained information at temporal level using only coarse annotation at the interview level. However, most of the studies restrict the analysis of attention mechanisms to the display of examples and do not conduct an in-depth analysis. Moreover, the validity of the attention curves as an explanation has recently been called into question \cite{Jain2019AttentionExplanation}.

\section{Experimental Setup}

\subsection{Dataset}
As our goal is to evaluate and assess the relevance of attention mechanisms that are already trained, we use the same database previously collected by us \cite{Hemamou2019HireNetInterviews}. This database contains real French asynchronous video interviews of 7938 candidates applying for 475 sales positions. Each interview of a specific position has the same number of questions predefined upstream by the recruiter. Once the candidates finish answering the set of predefined questions on the web platform, recruiters and managers can connect to this platform, watch these answers, and evaluate the candidate. They can like, dislike, shortlist candidates, evaluate them on predefined criteria or write comments. Based on this information, candidates who have been liked or shortlisted have been labelled "hirable", otherwise they are labelled "not hirable". If candidates received different annotations from multiple recruiters, a majority vote was taken. In case of draw, the candidate is considered "hirable". To the best of our knowledge, this database is the one with the highest number of real applicants assessed by real practitioners for a real position. We extracted verbal content using an automatic speech recognition tool (Google API). These Asynchronous Video Interviews have been recorded in the wild from various devices leading to a wide range of setups. Due to this condition, technical problems could occur such as videos without  audio, illumination problems in videos, or failure in the automatic speech recognition. Descriptive statistics for each modality are available in Table \ref{table:datasets}. The dataset can not be made available to the public due to high privacy constraints. 

\begin{table}
\centering
\begin{tabular}{|l|c|c|c|}
\hline
Modality                               & Text     & Audio  & Video  \\
\hline
\hline
Train set   & 6350     & 6034  & 5706  \\
\hline
Validation set & 794     & 754   & 687   \\
\hline
Test set   & 794     & 755   & 702   \\
\hline
\hline
\begin{tabular}[c]{@{}l@{}}Questions per\\ interview (mean)\end{tabular} & 5.05     & 5.10  & 5.01  \\
\hline
Total length                            & 3.82\,M\,words & 557.7\,h & 508.8\,h \\
\hline
\begin{tabular}[c]{@{}l@{}}Length per \\ question (mean)\end{tabular} & 95.2\,words  & 52.19\,s & 51.54\,s \\
\hline
\begin{tabular}[c]{@{}l@{}}\textit{Hirable} label\\ proportion\end{tabular} & 45.0\,\%     & 45.5\,\%  & 45.4\,\%  \\
\hline
\end{tabular}
\caption{Number of candidates in each set and overall statistics of the dataset.}
\label{table:datasets}
\end{table}

\subsection{\lhcorr{Our model}{HireNet}}
\label{HireNetPres}

In a previous article, we proposed  \lhcmt{HireNet,} an attention neural network to infer hirability from structured video interviews. \lhcorr{Our model}{HireNet} \cite{Hemamou2019HireNetInterviews} was conceived to represent a sequence of questions and their answers containing themselves a sequence of social signals. In the following sections, we focus only on the low level encoder of our model which aims to detect salient social signals. This encoder is a bidirectional Gated Recurrent Unit (GRU) \cite{cho2014learning} which encodes information from a sequence of low level descriptors. This encoder is followed by an attention mechanism that weights each timestep differently according to its importance. We aim to validate the usefulness of attention mechanisms to automatically extract the most useful slices to predict hirability. For the following study, we define an \textit{attention slice} as a slice selected according to the attention curve. For the sake of simplicity, we decided to focus only on visual features in the first question. The first question is highly linked to self presentation tactics, and initial impressions play an important role in the variance of interview scores \cite{Swider2016InitialOutcomes}. Moreover, our preliminary inspection of attention curves has shown that for the visual modality, attention peaks appear more frequently than for other modalities. These visual features are the position and orientation of the head, and continuous and categorical facial action units activations which have been extracted using OpenFace\cite{Baltrusaitis2018OpenFaceToolkit}. Values were smoothed with a time window of 0.5s and an overlap of 0.25s before being fed to our model. This duration is frequently used in the literature of Social Computing \cite{Varni2018ComputationalInteractions} and we validated it for our corpus by annotating the duration  of social signals in a set of videos.

We trained our model to achieve best results for the mean of F-1 score of the positive and negative class rather than only on the positive class as we did in \cite{Hemamou2019HireNetInterviews}. \lhcmt{We call this average \textit{Mean F1}}. \lhcorr{Moreover, for higher chances of obtaining quality attention curves, we chose to train five different models and then averaged the attention values}{As neural networks are subject to various variability sources such as random weights initialization, stochastic gradient descent or dropout, we chose to train five different instances of the model and then averaged the attention values. That way, we aim to capture the more general behaviour of attention mechanisms} \cite{Jetley2018LearnAttention}. Mean performance and confidence interval details on test set are reported in Table \ref{tab1ResultsHireNet}.

\begin{table}[htbp]
\caption{Performances of our model on test set for hirability prediction task}
\begin{center}
\begin{tabular}{|c|c|c|c|}
\hline
\textbf{Model}& F1 Positive Class& F1 Negative Class & \textit{Mean F1} \\
\hline
HireNet & 0.607 $\pm$ 0.023 & 0.628 $\pm$ 0.013 & 0.618 $\pm$ 0.008 \\
\hline
\end{tabular}
\label{tab1ResultsHireNet}
\end{center}
\end{table}

\section{ Do attention curves actually expose distinguishable peaks ?}
\label{Q1}

\subsection{Methodology : Extraction of \textit{attention slices} by unsupervised outlier detection}
\begin{figure}[htbp]
\centering
\centerline{\includegraphics[width=1.\columnwidth]{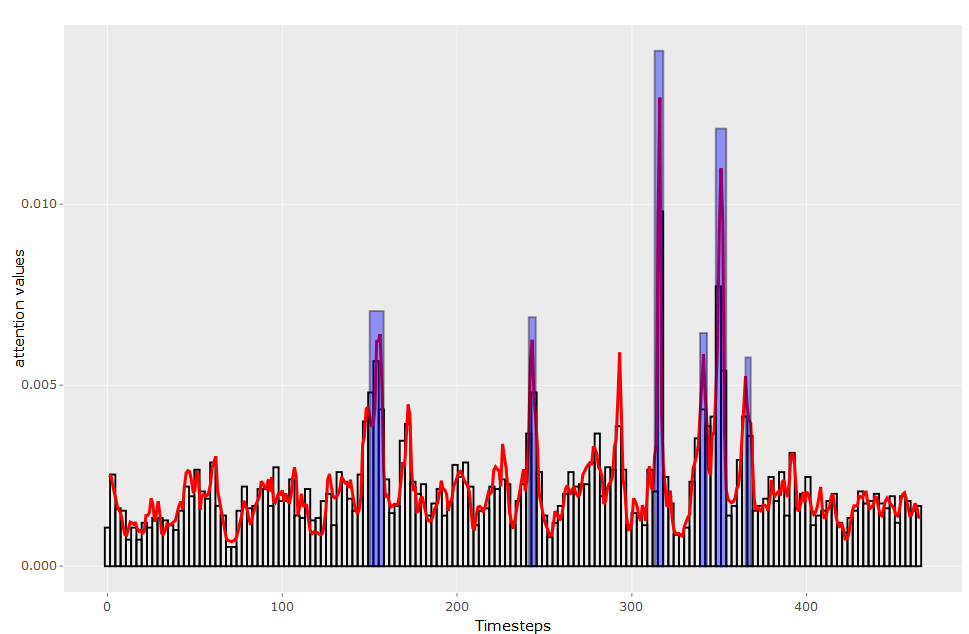}}
\captionsetup{justification=centering}
\caption{\textit{Attention slice} extraction.\\
The Attention curve is in red, the histogram of points drawn is the result of the sampling procedure, and the detected peaks are highlighted by blue boxes.}
\label{figPeaks}
\end{figure}

Attention curves mostly consist of noisy fluctuations with some high value peaks \cite{Yu2017TemporallyContent}\cite{Hemamou2019HireNetInterviews}. A typical example is the red curve in figure \ref{figPeaks}. The first step of our methodology consists of filtering attention curves containing peaks and then extracting where attention rises (\textit{attention slices}).
In order to achieve this, we use and adapt an unsupervised outlier detection method already proposed in \lhcmt{another study on attention} \cite{KimInterpretableAttention}.
We sample timescale by randomly selecting samples according to the distribution given by the attention curves. Then, points with higher attention values have a higher chance of being selected. An example of this sampling process is available in figure \ref{figPeaks}. 
Once this sampling is done, we use DBSCAN (Density-Based Spatial Clustering of Applications with Noise)\cite{Ester1996ANoise}, an unsupervised density based algorithm, which aims to find regions where the density of points drawn is higher. This method proved to be efficient because it manages the noisy values of attention curves, and the number and expansion of regions (duration in our special case of time series) do not need to be specified. A typical result of this algorithm is depicted by blue boxes in figure \ref{figPeaks}.

\subsection{Results and descriptive statistics about extracted peaks}

\begin{figure*}
\centering
\begin{minipage}{\columnwidth}
 \centering
 \includegraphics[width=0.8\columnwidth]{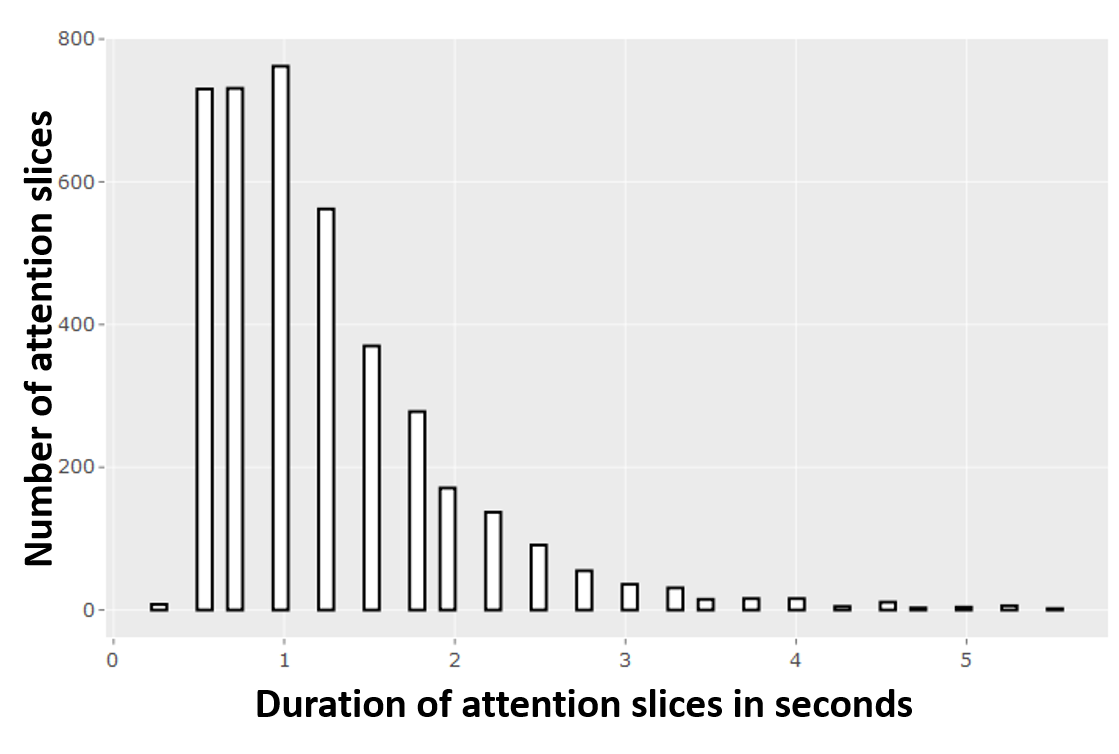}
 \caption{\lhcorr{Histogram of \textit{attention slices} depending on their duration in seconds.}{Histogram of the duration of \textit{attention slices}}}
 \label{fig:test1}
\end{minipage}
\begin{minipage}{\columnwidth}
 \centering
 \includegraphics[width=0.8\columnwidth]{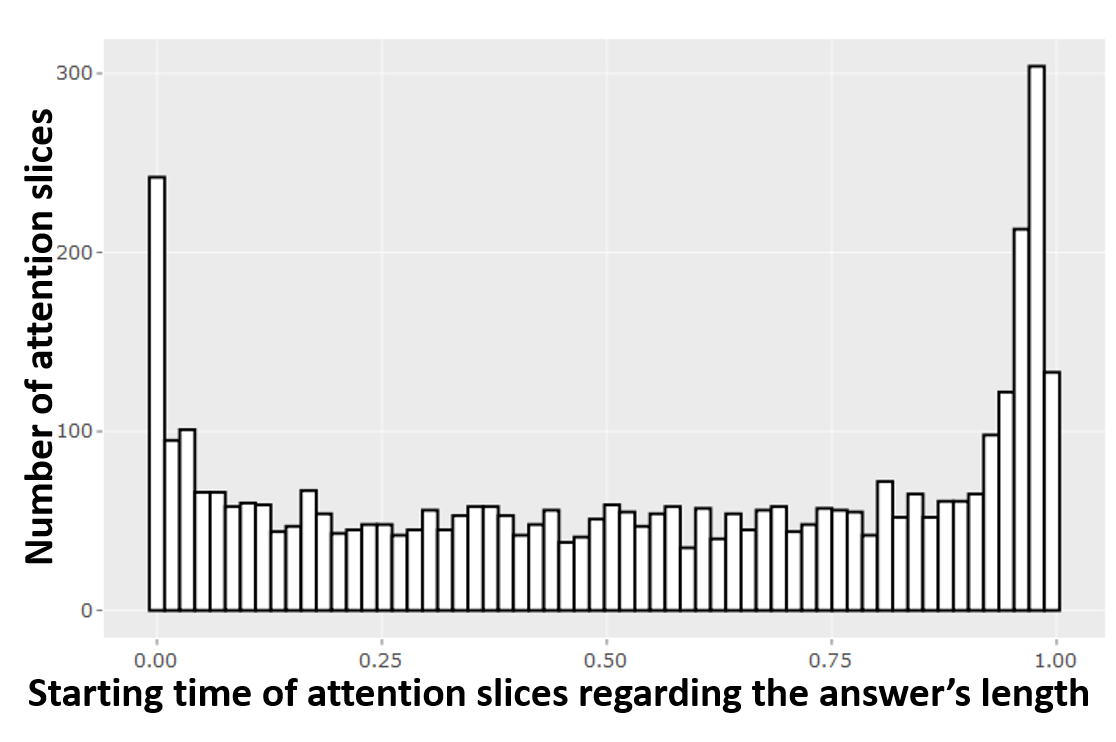}
 \caption{\lhcorr{Histogram of \textit{attention slices} depending on their starting time.}{Histogram of the starting time of \textit{attention slices} relative to the total duration of the answer}}
 \label{fig:test2}
\end{minipage}
\end{figure*}

\lhcorr{According to our procedure, some attention curves from candidate answers do not have distinguishable peaks. Consequently we lose a fraction of our dataset. Table \ref{tabSummary} provides a summary of the data serving as a basis for our study in terms of kept answers. 
We also give some numbers about how long and when the peak with the largest amplitude occurs during an interview respectively in figures \ref{fig:test1} and \ref{fig:test2}.}{Table \ref{tabSummary} provides a summary of the data serving as a basis for our study in terms of answers containing peaks. Some attention curves from candidate answers do not have peaks. In fact, some candidates may not display any particularly important moments during the answer to the first question. Figures \ref{fig:test1} and \ref{fig:test2} describe how long and when the peak with the largest amplitude occurs during an interview. }
It is interesting to note that the duration of the important slices extracted by the attention mechanism follows a very similar distribution to the duration of facial expressions which typically lasts between 0.5 and 4s\cite{Matsumoto2011EvidenceEmotion}. Moreover, it seems they occur more often at the beginning and at the end of an answer. Such cues could indicate that non verbal behaviours occurring at the beginning (turn taking) and at the end of the answer (turn giving) have a strong impact on \lhcorr{hirability decision}{recruiter's evaluation} as in \lhcmt{other} face to face interactions \cite{Cassell2001Non-verbalStructure,GOODRICH1979Face-to-FaceTheory}.

\begin{table}[htbp]
\caption{Descriptive table of number of answers containing peaks}
\begin{center}
\begin{tabular}{|c|c|}
\hline
\textbf{Set}  & \textbf{Percentage of answers containing peaks} \\
\hline
 Train and Validation & 63.8\% (3644 answers kept) \\
 \hline
 Test & 57.4\% (403 answers kept)\\
\hline

\end{tabular}
\label{tabSummary}
\end{center}
\end{table}

\section{Are social signals during attention slices different from those in random slices?}
\label{Q2}

\subsection{Method : Supervised classification between \textit{attention slices} and random slices}
\label{DistinguishAttention}
In this section we study the relationship between the values fed to our model and the \textit{attention slices}.
Attention values could heavily depend on the context and on the model's memory, and depend very little on the time-frame they point to. Our model uses Contextual Attention learned on top of a bidirectionnal GRU. GRU is a sequence modelling component that outputs vectors that depend on the current timestamp as well as it's previous output. We hypothesize that a rise in attention is due to a change of behaviour within an answer.

To ensure that attention mostly stems from what is happening in the concerned time-steps, we construct a binary classification task. For our task, we take as one class the most important \textit{attention slices} (the slice containing the peak with the highest amplitude) extracted in each candidate answer. As the other class, we take four moments with the same duration sampled in the candidate's answer according to a distribution proportional to ($1-attention\_values$). \lhcmt{ $attention\_values$ is the average of the output values of the attention mechanisms of the candidate's answer}. Through this sampling, we aim to select moments of varying importance, and not only the most unimportant ones, while still avoiding the most important moments. As our goal is to understand if \textit{attention slices} are different, we decided to use traditional classifiers with which a methodology to detect important features is well established. Thus, classifiers we used for this task are Lasso (linear and transparent model) and Random Forest (non linear model). As these classifiers take as input fixed vector, we summarize the features of the selected moments' time-windows through the use of the following functions: mean, mean of positive gradients and mean of negative gradients.

 We use these functions here as an attempt at capturing temporal dependencies while keeping an explainable set of features. As the attention mechanisms are trained on top of GRUs, they capture temporal variations. Our gradient functions have been used successfully in a previous behavioral classification work \cite{Ryoo2015PooledVideos}. These functions are applied on the same feature set used in subsection \ref{HireNetPres} to train HireNet, the unique difference remains on a preprocess step of Z-normalization regarding the whole answer. 
 For this experiment, we keep the same training, development, and test sets as we did before in \cite{Hemamou2019HireNetInterviews} to prevent any sort of data leaks.

\subsection{Results and analysis}

\begin{table}[htbp]
\caption{Classification results between random slices and \textit{attention slices}}
\begin{center}
\begin{tabular}{|c|c|c|c|}
\hline
\textbf{Model}& F1 Positive Class& F1 Negative Class & \textit{Mean F1} \\
\hline
\lhcmt{Random Baseline & 0.286 & 0.614 & 0.450} \\
Majority Class & 0 & 0.888 & 0.444 \\
Lasso & \allbold{0.812} & \allbold{0.955} & \allbold{0.884} \\
Random Forest & 0.760 & 0.945 & 0.852 \\
\hline
\end{tabular}
\label{tabDistinguish}
\end{center}
\end{table}

As shown in Table \ref{tabDistinguish} the classifier's performance is significantly above the random baseline, proving that despite the influence of sequence modelling and the use of context information, the importance of a moment is still mainly defined by the events occurring in it. It shows that specific moment where peaks of attentions occur are distinguishable from others slices of the same answer.

\subsection{Non verbal features importance analysis}

\begin{table*}[t]
\begin{center}
\caption{ \lhcmt{Feature Importance Analysis} }
\begin{tabular}{|p{3cm}|p{3.5cm}|p{3.5cm}|p{7.0cm}|}
\hline
\textbf{Group} & \multicolumn{2}{| c | }{ \textbf{Lasso} } &\textbf{Random Forest Permutation} \\
\cline{2-3} 
& Positive coefficients & Negative coefficients & \\
\hline
\hline
Lower Face & \bigbold{AU20$^{2}$}, \bigbold{AU23$^{4}$}, \bigbold{AU17$^{10}$} &\bigbold{AU26$^{3}$},AU12$^{6}$,\bigbold{$\suparrow$AU17$^{11}$}, \bigbold{$\suparrow$AU20$^{16}$},$\sdownarrow$AU10$^{19}$, \bigbold{AU25$^{20}$} & \bigbold{AU20$^{2}$}, \bigbold{AU26$^{3}$}, \bigbold{AU23$^{4}$} ,\bigbold{AU17$^{5}$}, \bigbold{AU25$^{6}$}, \bigbold{$\suparrow$AU23$^{9}$}, \bigbold{$\suparrow$AU17$^{14}$}, \bigbold{AU15$^{15}$}, AU12$^{16}$, \bigbold{AU14$^{20}$} \\
\hline
Upper Face & \bigbold{$\suparrow$AU7$^{5}$}, \bigbold{AU2$^{8}$}, \bigbold{AU4$^{12}$}, $\suparrow$AU2$^{14}$ , \bigbold{AU1$^{18}$} & $\sdownarrow$AU2$^{13}$, $\sdownarrow$ AU7$^{17}$ & \bigbold{AU4$^{7}$}, \bigbold{AU2$^{17}$} \\
\hline
Blink and Gaze & \bigbold{AU45$^{1}$} & $\sdownarrow$AU45$^{15}$  & \bigbold{AU45$^{1}$}, \bigbold{$\suparrow$AU45$^{12}$}, gaze\_angle$_y$ $^{18}$ \\
\hline
Position and rotation of the head & \bigbold{ $\suparrow$ $T_z^{7}$} , \bigbold{ $\sdownarrow$ $T_z^{9}$ } &  &  $T_z^{10}$,  $Ty^{11}$, $Rx^{13} $\\
\hline
Confidence of OpenFace & & & \bigbold{confidence$^{8}$}, \bigbold{$\sdownarrow$confidence$^{19}$} \\ 
\hline
\end{tabular}
$F^{i}$ denotes that feature $F$ is ranked in $i_{th}$ position. $\suparrow$ and $\sdownarrow$ stand respectively for mean of positive and negative gradients. Bold indicates that feature $F$ is significantly different from random slices on the test set based on two tailed t-test $p<0.0001$ 

\label{tabClassesImp}
\end{center}
\end{table*}

An experiment about features' importance has also been done in order to highlight the features that contribute the most when identifying \textit{attention slices}. Such analysis provides useful knowledge about what lets a slice be selected.
In order to obtain this feature analysis, we inspect coefficients of Lasso Model and rank them according to their magnitude. Concerning, Random Forest importance, we run a permutation importance analysis through the use of Boruta Package \cite{Kursa2010FeaturePackage}. 
The result table for top twenty features of both methods is available in table \ref{tabClassesImp}.
Blinking (AU45), lip stretcher (AU20), jaw drop (AU26) and lip tightener (AU23) are considered by both analyses to be the top 4 features with the most importance. Based on the sign of the coefficients of the Lasso model, we notice that \textit{attention slices} are induced by: i) Eyes closed longer than usual, ii) The activation of lip stretcher and lip tightener, iii) The non-activation of jaw drop.
These cues could indicate that moments when a candidate is not talking (absence of jaw drop) or when he/she displays social cues of anxiety (lip stretcher and lip tightener) are considered more important by the attention mechanism\cite{Feiler2016BehavioralAnxiety}. Frames with chin raiser (AU17) are also considered important.
Also, outer brow raiser (AU2) and brow lowerer (AU4) appear in both features importance analyses. Coefficients of Lasso and directions of gradients support that moments when candidates raise and keep outer brow raised are also judged more important. 
Another interesting features is the use of the depth position ($T_z$). This analysis seems to indicate that movements back and forth could also be detected as important moments.
Finally it's interesting to highlight the confidence of OpenFace and negative gradients of OpenFace's confidence are selected by Random Forest permutation analysis. 

\section{Are attention slices more informative with regard to hirability than random slices ?}
\label{Q3}

\begin{table*}[bpht]
\caption{Result of classification task for hirability prediction from \textit{attention slices} and random slices}
\begin{center}
\begin{tabular}{|c|c|c|c|c|}
\hline
\textbf{Thin slices integrated}&\multicolumn{4}{|c|}{\textbf{AUC}} \\
\cline{2-5} 
\textbf{} & \textbf{\textit{Random Forest*}}& \textbf{\textit{Lasso*}}& \textbf{\textit{SVM Linear*}} & \textbf{\textit{SVM RBF*}} \\
\hline
Random thin slices & 0.545 $\pm$ 0.005 & 0.517 $\pm$ 0.005 & 0.518 $\pm$ \lhcorr 0.005  & 0.528 $\pm$ 0.005\\
\textit{Attention slices} & \allbold{0.554 $\pm$ 0.003} & \allbold{0.550 $\pm$ 0.003} & \allbold{ 0.543 $\pm$ 0.004 } & \allbold{0.537  $\pm$ 0.003} \\
\hline
\multicolumn{5}{l}{}
\\ [-1.75ex]
\multicolumn{5}{c}{\lhcmt{statistical significance is based on two-tailed t-test \textsuperscript{*}$p<0.01$}}
\end{tabular}
\label{tabDistinguishv2}
\end{center}
\end{table*}


\subsection{Method : Supervised classification of hirability based on random slices or attention slices}
\lhcorr{So far, we established that our model could emphasize a number of moments in a candidate's answer, and found that these \textit{attention slices} are different from other selected slices. But we still did not validate the \textit{attention slices} as useful for hirability prediction task.}{} We intend to evaluate that the moments highlighted by the attention mechanism carry more useful information than random moments. The following procedure aims at testing that the highlighted moments have superior predictive capabilities compared to the rest of the interview.

\lhcorr{We constructed a classification task based on a minimal time-window in a candidate's answer}{We constructed a classification task based on only one slice of the candidate’s answer}. We ran two instances of this task: The first one uses the most important moment for each candidate as judged by the attention mechanism, while the second one uses the same sampling as in section \ref{DistinguishAttention}. We used the same features as in section \ref{DistinguishAttention}.
As the input slices are different from the ones required by our model (classification of structured asynchronous video interview), we choose to experiment with non-sequential algorithms. We divide our algorithms into 3 sets. The first set is only composed of Lasso, an L1 regularized linear classifier trained with the same loss as HireNet: a binary cross-entropy. This first set has processing capabilities inferior to those of HireNet. In fact, inspite of their complexity, the GRUs composing our hierachical model process each input with only one non-linearity. Consequently, it is capable of drawing 3 linear separators: 1 for each hierarchy level, and 1 for the final dense layer, and of adding up sequential elements in a learnable fashion. The second set is comprised of only linear-SVM. It is an algorithm with strictly inferior processing capabilities compared to HireNet, trained with a different loss function. The third set comprises SVM with a Radial Basis Function kernel (RBF) and Random Forest, two algorithms with processing capabilites unavailable in HireNet, and with loss functions different than that of HireNet. We choose the Area Under the Curve (AUC) as evaluation metric, as it has the advantage of not requiring any threshold and it is suited to comparing different models. 
For each of the algorithms used, we performed a bootstrapping procedure as follows; we trained 100 instances, each on a subset of the training set sampled with replacement. We then obtained a set of scores that allowed us to calculate confidence intervals for our results to get a sense of their statistical significance.

\subsection{Results and discussion.}

Results are reported in Table \ref{tabDistinguishv2}. \lhcmt{We observed only based on short slices of 0.5s to 4s, that the prediction of hirability is above random.}
For \lhcorr{Lasso}{all the classifiers}, the results show statistically significant differences in the predictive performance of the \textit{attention slices} in comparison to random slices.
\lhcorr{For the remaining classifiers, the ones fed with \textit{attention slices} achieve better performance with a smaller variance. However, none of the differences between the sets of slices are statistically significant}{We can note that the importance of the use of attention slices is clearer (larger performance gap) for linear classifiers compared to non linear classifiers}. This highlights an important consideration to keep in mind when using attention mechanisms: The \textit{attention slices} are selected with regard to the learner they are fed to. As shown by our results, \lhcorr{both processing capabilities and training loss function vary the obtained importance of thin slices}{processing capabilities vary the obtained importance of thin slices}.

\section{Conclusion, Limits and Future works}

In this paper, we established that moments with peaks of attention are different from randomly picked slices. We described this difference in terms of the input visual social signals. Visual cues seem to relate with anxiety (activation of lip strecher and lip tightener), blinking and pauses (non-activation of jaw drop). \lhcorr{Concerning the confidence of OpenFace,  after watching some videos where this drop of confidence occurs, we can notice that it's often resulting from face occlusion and mostly from self-touching, a social signal that could be interesting to study in the context of asynchronous video interviews.}{} \textit{Attention slices} are more likely to occur during turn taking (at the beginning of the answer) and turn giving (at the end of the answer) as in real face to face interaction \cite{GOODRICH1979Face-to-FaceTheory,Cassell2001Non-verbalStructure}. We also study the predictive value of the selected moments in comparison to random moments, and consequently put into perspective the use of the expression "important moment" to qualify an interview slice.
In future work, we would like to investigate different types of attention mechanisms \cite{Martins2016FromClassification} in our hirability prediction model, and a larger range of classifiers for the study of \textit{attention slices}. \lhcorr{As attention values distinguish only importance and not positive or negative influence, We would also like to propose a methodology that distinguishes attention slices that have a positive impact from ones having a negative impact on perceived hirability. Besides, as a first step, we limited our study to the first question of the interview, to the visual features and only to the slices with the highest attention value.}{Attention values highlight the importance of moments ,but do not include information about whether they have a positive or negative impact on recruiters' decisions. We aim in our future work to distinguish attention slices that have a positive impact from ones having a negative impact.} We plan to expand our work to other questions, modalities and the use of more than the most important peaks. \lhcmt{ Finally, as our approach is based on a learned model (eg HireNet), one research direction is to improve it in terms of performance and bias control.} Next steps of our work will also be dedicated to the design of a procedure to quantify the proportion of important moments. In that sense, we plan to conduct an annotation task and a user study, in order to: \textit{i)} quantify the aforementioned proportion; \textit{ii)} study links between macro cues and micro cues under the light of \textit{attention slices} spotted by our model; \textit{iii)} build an interface that provides useful feedback for candidates, and higher decision transparency for recruiters.

\section*{Acknowledgment}

This work was supported by the company EASYRECRUE. We would like to thank Jeremy Langlais and Amandine Reitz for their support and their help. We would also like to thank  Erin Douglas for proofreading the article.

\bibliographystyle{plain}
\bibliography{references.bib}

\begin{thebibliography}{10}

\bibitem{Baltrusaitis2018OpenFaceToolkit}
Tadas Baltrusaitis, Amir Zadeh, Yao~Chong Lim, and Louis~Philippe Morency.
\newblock {OpenFace 2.0: Facial behavior analysis toolkit}.
\newblock {\em Proceedings - 13th IEEE International Conference on Automatic
  Face and Gesture Recognition, FG 2018}, pages 59--66, 2018.

\bibitem{Carney2007AImpressions}
Dana~R. Carney, C.~Randall Colvin, and Judith~A. Hall.
\newblock {A thin slice perspective on the accuracy of first impressions}.
\newblock {\em Journal of Research in Personality}, 41(5):1054--1072, 10 2007.

\bibitem{Cassell2001Non-verbalStructure}
Justine Cassell, Yukiko~I Nakano, Timothy~W Bickmore, Candace~L Sidner, and
  Charles Rich.
\newblock {Non-verbal cues for discourse structure}.
\newblock In {\em Proceedings of the 39th Annual Meeting on Association for
  Computational Linguistics - ACL '01}, pages 114--123, Morristown, NJ, USA,
  2001. Association for Computational Linguistics.

\bibitem{Chen2017}
Lei Chen, Ru~Zhao, Chee~Wee Leong, Blair Lehman, Gary Feng, and Mohammed~Ehsan
  Hoque.
\newblock {Automated video interview judgment on a large-sized corpus collected
  online}.
\newblock In {\em 2017 Seventh International Conference on Affective Computing
  and Intelligent Interaction (ACII)}, pages 504--509. IEEE, 10 2017.

\bibitem{cho2014learning}
Kyunghyun Cho, Bart Van~Merri{\"{e}}nboer, Caglar Gulcehre, Dzmitry Bahdanau,
  Fethi Bougares, Holger Schwenk, and Yoshua Bengio.
\newblock {Learning phrase representations using RNN encoder-decoder for
  statistical machine translation}.
\newblock {\em arXiv preprint arXiv:1406.1078}, 2014.

\bibitem{Chollet2017AssessingBehavior}
Mathieu Chollet and Stefan Scherer.
\newblock {Assessing Public Speaking Ability from Thin Slices of Behavior}.
\newblock {\em Proceedings - 12th IEEE International Conference on Automatic
  Face and Gesture Recognition, FG 2017 - 1st International Workshop on
  Adaptive Shot Learning for Gesture Understanding and Production, ASL4GUP
  2017, Biometrics in the Wild, Bwild 2017, Heteroge}, pages 310--316, 2017.

\bibitem{Degroot2009CanInterviews}
Timothy Degroot and Janaki Gooty.
\newblock {Can nonverbal cues be used to make meaningful personality
  attributions in employment interviews?}
\newblock {\em Journal of Business and Psychology}, 24(2):179--192, 2009.

\bibitem{Ester1996ANoise}
Martin Ester, Hans-peter Kriegel, Xiaowei Xu, and D~Miinchen.
\newblock {A Density-Based Algorithm for Discovering Clusters in Large Spatial
  Databases with Noise}.
\newblock 1996.

\bibitem{Eyben2016TheComputing}
Florian Eyben, Klaus~R. Scherer, Bjorn~W. Schuller, Johan Sundberg, Elisabeth
  Andre, Carlos Busso, Laurence~Y. Devillers, Julien Epps, Petri Laukka,
  Shrikanth~S. Narayanan, and Khiet~P. Truong.
\newblock {The Geneva Minimalistic Acoustic Parameter Set (GeMAPS) for Voice
  Research and Affective Computing}.
\newblock {\em IEEE Transactions on Affective Computing}, 7(2):190--202, 2016.

\bibitem{Feiler2016BehavioralAnxiety}
Amanda~R. Feiler and Deborah~M. Powell.
\newblock {Behavioral Expression of Job Interview Anxiety}.
\newblock {\em Journal of Business and Psychology}, 31(1):155--171, 2016.

\bibitem{FORBES1980NonverbalInterviews}
Ray~J. Forbes and Paul~R. Jackson.
\newblock {Non‐verbal behaviour and the outcome of selection interviews}.
\newblock {\em Journal of Occupational Psychology}, 53(1):65--72, 1980.

\bibitem{Frauendorfer2015TheInterview}
Denise Frauendorfer and Marianne~Schmid Mast.
\newblock {The Impact of Nonverbal Behavior in the Job Interview}.
\newblock In {\em The Social Psychology of Nonverbal Communication}, pages
  220--247. Palgrave Macmillan UK, London, 2015.

\bibitem{Gifford1985NonverbalJudgments.}
Robert Gifford, Cheuk~Fan Ng, and Margaret Wilkinson.
\newblock {Nonverbal cues in the employment interview: Links between applicant
  qualities and interviewer judgments.}
\newblock {\em Journal of Applied Psychology}, 70(4):729--736, 1985.

\bibitem{GOODRICH1979Face-to-FaceTheory}
Wells Goodrich.
\newblock {Face-to-Face Interaction: Research, Methods, and Theory}.
\newblock {\em Family Process}, 18(3):355--356, 9 1979.

\bibitem{Hemamou2019HireNetInterviews}
Leo Hemamou, Ghazi Felhi, Vincent Vandenbussche, Jean-claude Martin, and Chloe
  Clavel.
\newblock {HireNet : a Hierarchical Attention Model for the Automatic Analysis
  of Asynchronous Video Job Interviews}.
\newblock In {\em AAAI}, 2019.

\bibitem{Hoque2013MACHb}
Mohammed~Ehsan Hoque, Matthieu Courgeon, Jean-Claude Martin, Bilge Mutlu, and
  Rosalind~W. Picard.
\newblock {MACH}.
\newblock In {\em Proceedings of the 2013 ACM international joint conference on
  Pervasive and ubiquitous computing - UbiComp '13}, page 697, New York, New
  York, USA, 2013. ACM Press.

\bibitem{Jain2019AttentionExplanation}
Sarthak Jain and Byron~C. Wallace.
\newblock {Attention is not Explanation}.
\newblock In {\em North American Chapter of the Association for Computational
  Linguistics}, 2019.

\bibitem{Jetley2018LearnAttention}
Saumya Jetley, Nicholas~A Lord, Namhoon Lee, and Philip H~S Torr.
\newblock {Learn To Pay Attention}.
\newblock In {\em International Conference on Learning Representations}, pages
  1--14, 4 2018.

\bibitem{KimInterpretableAttention}
Jinkyu Kim and John Canny.
\newblock {Interpretable Learning for Self-Driving Cars by Visualizing Causal
  Attention}.

\bibitem{Kursa2010FeaturePackage}
Miron~B Kursa.
\newblock {Feature Selection with the Boruta Package}.
\newblock {\em Journal of Statistics}, 36(11):1--13, 2010.

\bibitem{Liu2019ImprovingDistillation}
Xuan Liu, Xiaoguang Wang, and Stan Matwin.
\newblock {Improving the interpretability of deep neural networks with
  knowledge distillation}.
\newblock {\em IEEE International Conference on Data Mining Workshops, ICDMW},
  2018-Novem:905--912, 2019.

\bibitem{Martins2016FromClassification}
André F.~T. Martins and Ramón~Fernandez Astudillo.
\newblock {From Softmax to Sparsemax: A Sparse Model of Attention and
  Multi-Label Classification}.
\newblock In {\em International Conference on Machine Learning}, 2016.

\bibitem{Matsumoto2011EvidenceEmotion}
David Matsumoto and Hyi~Sung Hwang.
\newblock {Evidence for training the ability to read microexpressions of
  emotion}.
\newblock {\em Motivation and Emotion}, 35(2):181--191, 2011.

\bibitem{Murphy2015ReliabilityInteractions}
Nora~A. Murphy, Judith~A. Hall, Marianne Schmid~Mast, Mollie~A. Ruben, Denise
  Frauendorfer, Danielle Blanch-Hartigan, Debra~L. Roter, and Laurent Nguyen.
\newblock {Reliability and Validity of Nonverbal Thin Slices in Social
  Interactions}.
\newblock {\em Personality and Social Psychology Bulletin}, 41(2):199--213, 2
  2015.

\bibitem{Naim2018AutomatedPerformance}
Iftekhar Naim, Md.~Iftekhar Tanveer, Daniel Gildea, and Mohammed~Ehsan Hoque.
\newblock {Automated Analysis and Prediction of Job Interview Performance}.
\newblock {\em IEEE Transactions on Affective Computing}, 9(2):191--204, 4
  2018.

\bibitem{Nambiar2017AutomaticInterviews}
Shruthi~Kukal Nambiar, Rahul Das, Sowmya Rasipuram, and Dinesh~Babu Jayagopi.
\newblock {Automatic generation of actionable feedback towards improving social
  competency in job interviews}.
\newblock In {\em Proceedings of the 1st ACM SIGCHI International Workshop on
  Multimodal Interaction for Education - MIE 2017}, pages 53--59, New York, New
  York, USA, 2017. ACM Press.

\bibitem{Nguyen2014HireBehavior}
Laurent~Son Nguyen, Denise Frauendorfer, Marianne~Schmid Mast, and Daniel
  Gatica-Perez.
\newblock {Hire me: Computational Inference of Hirability in Employment
  Interviews Based on Nonverbal Behavior}.
\newblock {\em IEEE Transactions on Multimedia}, 16(4):1018--1031, 6 2014.

\bibitem{Nguyen2015IMinute}
Laurent~Son Nguyen and Daniel Gatica-Perez.
\newblock {I Would Hire You in a Minute}.
\newblock In {\em Proceedings of the 2015 ACM on International Conference on
  Multimodal Interaction - ICMI '15}, pages 51--58, New York, New York, USA,
  2015. ACM Press.

\bibitem{RaoS.B2017AutomaticStudy}
Pooja Rao S.~B, Sowmya Rasipuram, Rahul Das, and Dinesh~Babu Jayagopi.
\newblock {Automatic assessment of communication skill in non-conventional
  interview settings: a comparative study}.
\newblock In {\em Proceedings of the 19th ACM International Conference on
  Multimodal Interaction - ICMI 2017}, number November, pages 221--229, New
  York, New York, USA, 2017. ACM Press.

\bibitem{Ribeiro2016WhyClassifier}
Marco Ribeiro, Sameer Singh, and Carlos Guestrin.
\newblock {“Why Should I Trust You?”: Explaining the Predictions of Any
  Classifier}.
\newblock In {\em Proceedings of the 2016 Conference of the North American
  Chapter of the Association for Computational Linguistics: Demonstrations},
  pages 97--101, Stroudsburg, PA, USA, 2016. Association for Computational
  Linguistics.

\bibitem{Ryoo2015PooledVideos}
M.~S. Ryoo, Brandon Rothrock, and Larry Matthies.
\newblock {Pooled motion features for first-person videos}.
\newblock {\em Proceedings of the IEEE Computer Society Conference on Computer
  Vision and Pattern Recognition}, 07-12-June(Figure 1):896--904, 2015.

\bibitem{2012TheSelection}
Neal Schmitt, editor.
\newblock {\em {The Oxford Handbook of Personnel Assessment and Selection}}.
\newblock Oxford University Press, 3 2012.

\bibitem{Schneider2015CuesInterview}
Leann Schneider, Deborah~M. Powell, and Nicolas Roulin.
\newblock {Cues to deception in the employment interview}.
\newblock {\em International Journal of Selection and Assessment},
  23(2):182--190, 2015.

\bibitem{Sundararajan2017AxiomaticNetworks}
Mukund Sundararajan, Ankur Taly, and Qiqi Yan.
\newblock {Axiomatic Attribution for Deep Networks}.
\newblock In {\em International Conference on Machine Learning}, 2017.

\bibitem{Swider2016InitialOutcomes}
Brian~W. Swider, Murray~R. Barrick, and T.~Brad~Harris.
\newblock {Initial impressions: What they are, what they are not, and how they
  influence structured interview outcomes}.
\newblock {\em Journal of Applied Psychology}, 101(5):625--638, 2016.

\bibitem{Varni2018ComputationalInteractions}
Giovanna Varni, Isabelle Hupont, Chloe Clavel, and Mohamed Chetouani.
\newblock {Computational Study of Primitive Emotional Contagion in Dyadic
  Interactions}.
\newblock {\em IEEE Transactions on Affective Computing}, 3045(c):1--1, 2018.

\bibitem{Wortwein2015MultimodalAssessment}
Torsten W{\"{o}}rtwein, Mathieu Chollet, Boris Schauerte, Louis-Philippe
  Morency, Rainer Stiefelhagen, and Stefan Scherer.
\newblock {Multimodal Public Speaking Performance Assessment}.
\newblock {\em International Conference on Multimodal Interaction (ICMI)},
  (March 2016):43--50, 2015.

\bibitem{Yosinski2015UnderstandingVisualization}
Jason Yosinski, Jeff Clune, Anh Nguyen, Thomas Fuchs, and Hod Lipson.
\newblock {Understanding Neural Networks Through Deep Visualization}.
\newblock In {\em In ICML Workshop on Deep Learning}, 6 2015.

\bibitem{Yu2017TemporallyContent}
Hongliang Yu, Liangke Gui, Michael Madaio, Amy Ogan, Justine Cassell, and
  Louis-Philippe Morency.
\newblock {Temporally Selective Attention Model for Social and Affective State
  Recognition in Multimedia Content}.
\newblock {\em Proceedings of the 2017 ACM on Multimedia Conference}, pages
  1743--1751, 2017.

\end{thebibliography}

\end{document}